\documentclass[12pt,journal]{article}
\usepackage{times,latexsym}
\usepackage{url}
\usepackage[T1]{fontenc}

\usepackage{amsmath}
\usepackage{amsfonts}
\usepackage{amssymb}
\usepackage{natbib}
\usepackage{algorithm}
\usepackage{algpseudocode}
\usepackage{amsthm}
\usepackage{dsfont}
\usepackage{tabularray}
\usepackage{bbm}
\usepackage{dsfont}
\usepackage{graphicx}
\usepackage{mathtools}

\numberwithin{equation}{section}

\begin{document}

\title{
Semantic Operator Prediction and Applications
}
\author{Farshad Noravesh\footnote{Email: noraveshfarshad@gmail.com}}
\maketitle


\begin{abstract}
In the present paper, semantic parsing challenges are briefly introduced and QDMR formalism in semantic parsing is implemented using sequence to sequence model with attention but uses only part of speech(POS) as a representation of words of a sentence to make the training as simple and as fast as possible and also avoiding curse of dimensionality as well as overfitting. It is shown how semantic operator prediction could be augmented with other models like the CopyNet model or the recursive neural net model. \\
\end{abstract}

\section{Introduction}
Semantic parsing and question answering have become coupled in recent years due to many reasons such as a technical reason, namely distant supervision, since creating a dataset for question answering pairs are much simpler than treebanks. Another type of weak supervision is to consider logical form of highest node in the tree as the only source of supervision as is done in \citep{Herzig2021}.

State of the art models to knowledge based question answering(KBQA) is observed to be based on semantic parsing to produce logical forms that can be easily executed on these knowledge graphs as is mentioned in \citep{Gu2022} , \citep{Gu2021},\citep{Berant2013} or separating semantic parsing task from the knowledge base interaction which is proposed in \citep{Ravishankar2021}.

Traditionally, entity linking(finding entities mentioned in the given question) has been considered as subproblem of semantic parsing and it is assumed that it is done beforehand, while \citep{Krishnamurthy2017} has combined entity linking with semantic parsing.
There are many formalisms in semantic parsing like abstract meaning representation(AMR), discourse representation structure(DRS), structured query language (e.g., SQL), and  lambda calculus. \citep{Kapanipathi2020} uses AMR for KBQA which integrates multiple, reusable modules like semantic parser, entity and relationship linkers, and neuro-symbolic reasoner. Instead of KBQA, the source of data could be based on tables. In this category, the answer format could be short term entity. It could also be a free form text like the FeTaQA dataset in \citep{Linyong2022} .

Semantic parsing is the building block of many challenging problems in artificial intelligence such as dialogue systems, question answering , and enhances technologies based on conversational AI, reading comprehension, and story generation. Traditional approaches to question answering  such as \citep{Zhou2018},\citep{Shi2021},\citep{Zhang2022},\citep{Ren2021} do not leverage semantic parsing and therefore their approaches are less explainable and interpretable and is hard to generalize. This is even harder for open domain question answering such as \citep{Sun2018},\citep{Sun2019}. There are three approaches to parsing in general, namely top-to-bottom, bottom-up, hybrid.
Although the bottom up constituency parsing introduced in \citep{Yang2022} is efficient but it is not scalable since getting these complex sequence annotations is expensive from crowdsourcing perspective. This suggests two paradigms to handle this problem. The first idea is to use distant supervision such that the error from question answering problem is backpropagated down to semantic parsing. The second idea which is proposed by  \citep{Wolfson2020} creates a middle layer to make the crowdsourcing cheaper and more scalable. 
Section~\ref{ESP} demonstrates why some approaches are not scalable and are expensive to be implemented in practice. Section~\ref{QD} shows how QDMR formalism is helpful for any scalable algorithm for semantic parsing. In section~\ref{model} a model is suggested. One of the main contributions of the present paper is emphasizing on lexicon-style alignments and disentangled information processing. In recent years, there has been interest on leveraging semantic tagging for semantic parsing as is done in \citep{Zheng2020} by first seeing semantic tags as latent variables and then using these semantic tags sequence to learn the logical form like either SQL type or lambda calculus. The training can be done either separately or jointly in an End-to-End way.
\section{Expensive Semantic Parsing\label{ESP}}
\begin{figure} [h!]
\centering
  \includegraphics[width=130mm,scale=0.5]{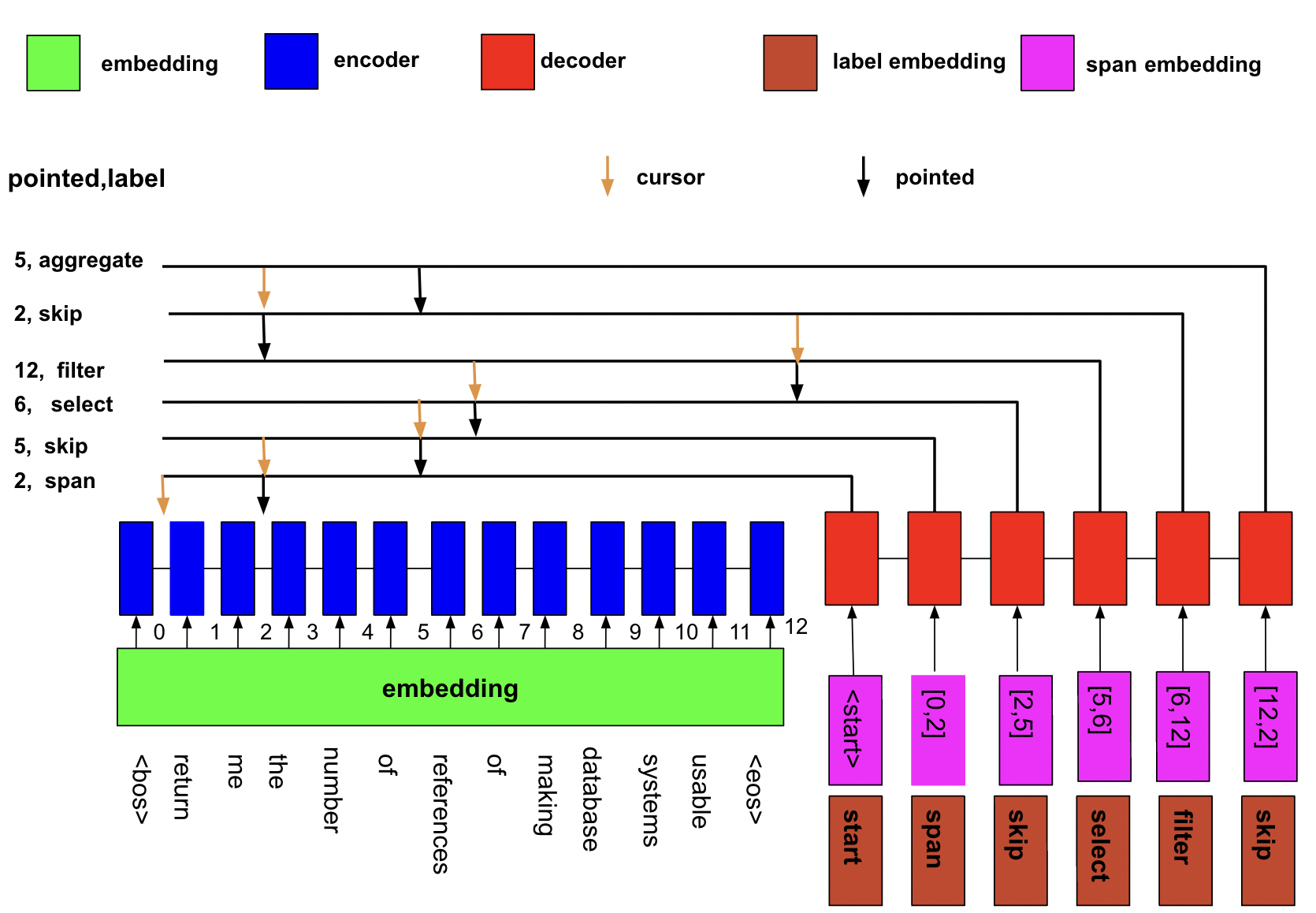}
  \caption{expensive model}
  \label{fig:expensivemodel} 
\end{figure}

Creating treebank is the first major challenge of current semantic parsing methods. This issue becomes even more dramatic than creating treebanks in syntactic parsing, since apart from ambiguities, the crowdsourcing agents are more expensive and each sentence takes more time to be annotated and annotators should be familiar with complex formalisms like combinatory categorial grammar(CCG),lambda calculus, type raising and composition in combinatory categorial grammar (CCG).   
\begin{figure} [h!]
\centering
  \includegraphics[width=130mm,scale=0.5]{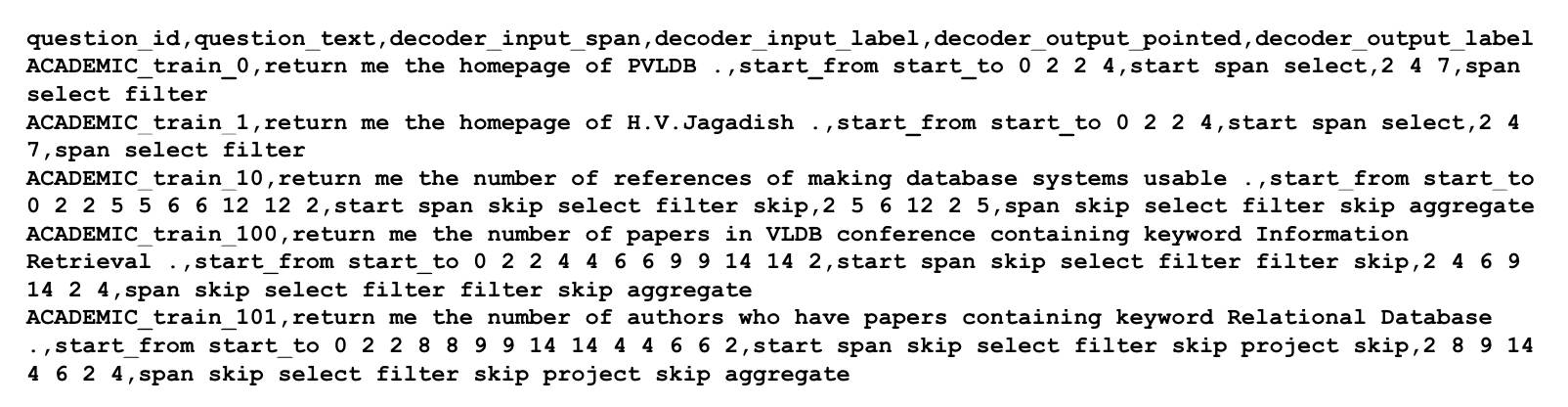}
  \caption{expensive dataset}
  \label{fig:expensive} 
\end{figure}
Using pointer network as is done in  \citep{Yang2022} needs expensive crowdsourcing. Although it was used for syntactic parsing, one can use the same methodology and apply it to semantic parsing by creating a dataset like figure~\ref{fig:expensive} which is very expensive in practice. Such an imaginary expensive model is shown in figure~\ref{fig:expensivemodel}. 

\begin{figure} [h!]
\centering
  \includegraphics[width=130mm,scale=0.5]{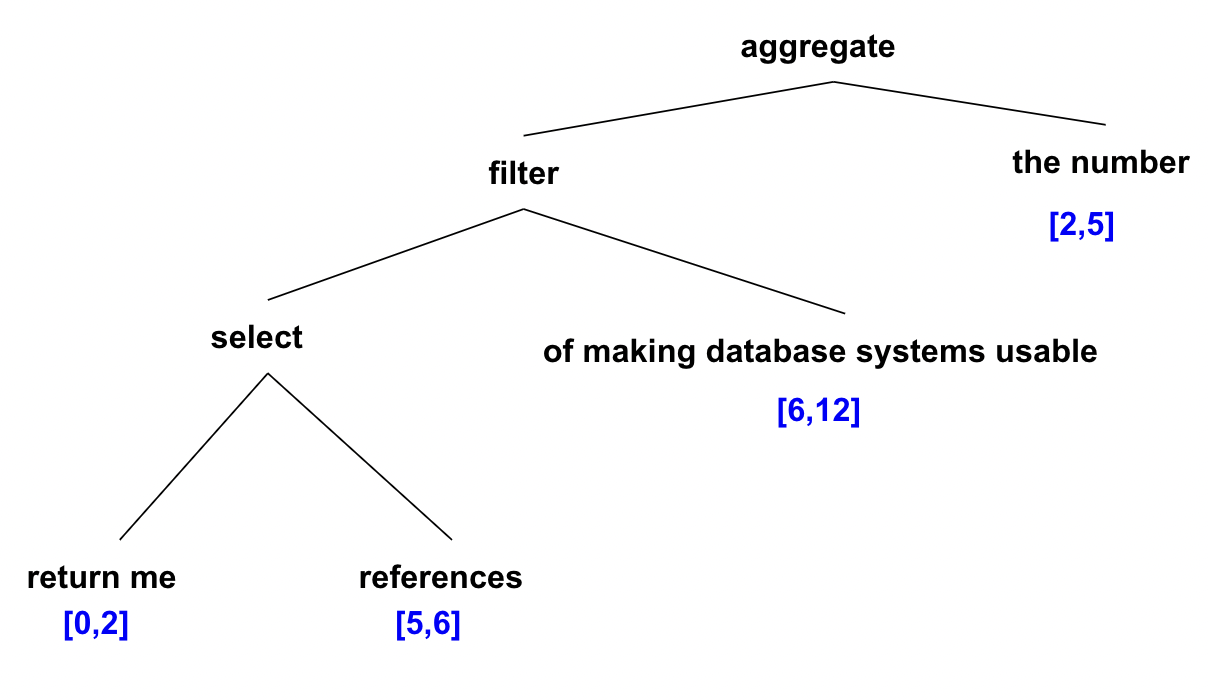}
  \caption{expensive parsing}
  \label{fig:expensiveparsing} 
\end{figure}

A typical sentence in QDMR dataset \citep{Wolfson2020} can be parsed using semantic operators as is shown in figure~\ref{fig:expensiveparsing}. For example, when the decoder is at cursor 6, it points to boundary 12(of making database systems usable)
and semantic operator label for it is "filter" . 

Recently, semantic parsing modeling is being done in stages like \citep{Dong2018} which handles the input utterance in some steps ranging from coarse level to fine details. Thus, they first generate a rough sketch of its meaning, where variable names and arguments is
glossed over. Then, missing details in sketch itself is filled in appropriately by the details inside the input utterance.
Another example of staging is \citep{Wolfson2020} which tackles the problem from a different perspective by creating a middle layer that is much easier to annotate for crowdsourcing and does not need any expertise in complex logical forms, lambda calculus and CCG. The next section shows how this staging mechanism in BREAK dataset could accelerate annotation process and create a big dataset relatively cheaply.

\section{Question Decomposition\label{QD}}
\begin{figure} [h!]
\centering
  \includegraphics[width=100mm,scale=0.5]{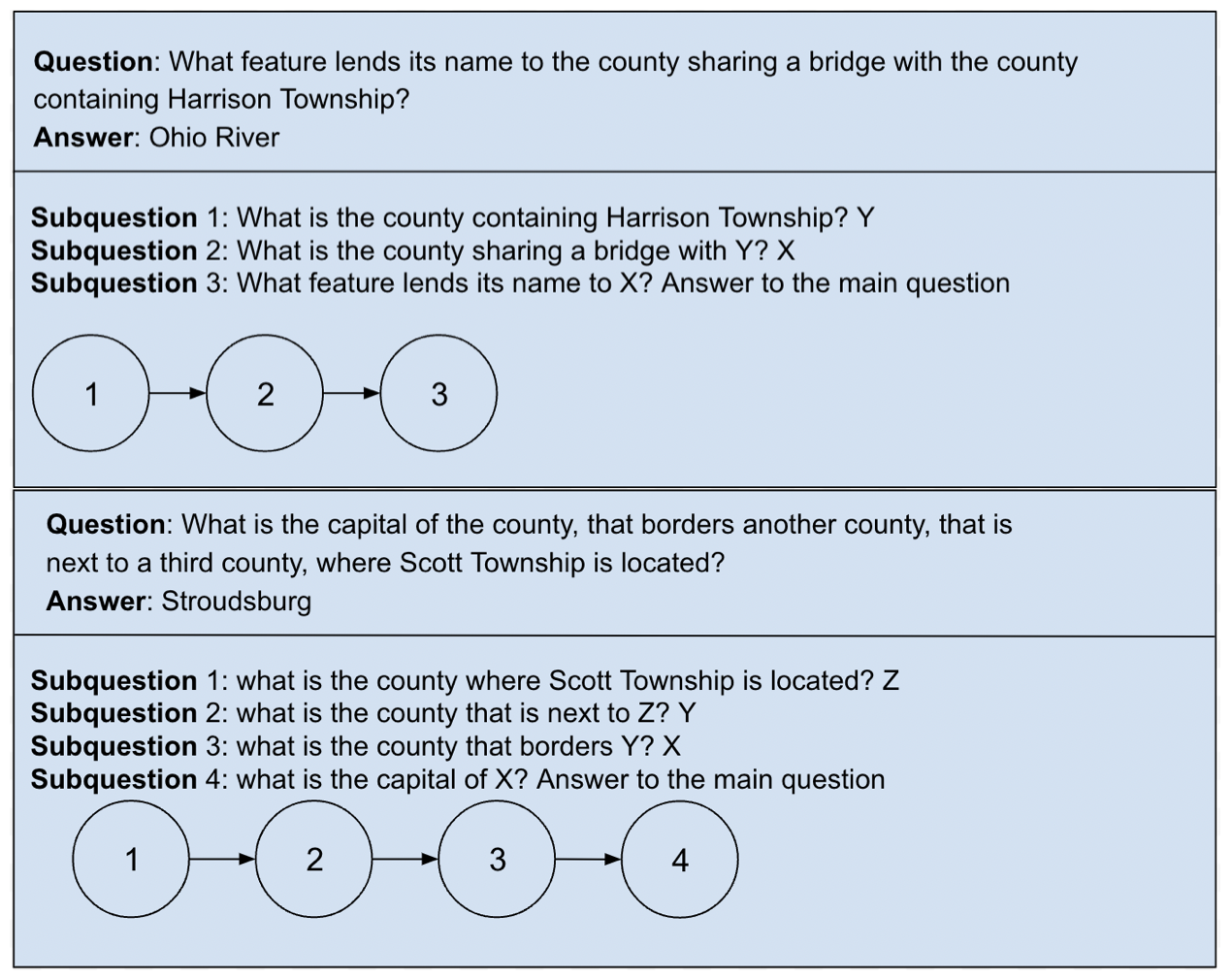}
  \caption{question decomposition}
  \label{fig:questions} 
\end{figure}

\begin{figure} [h!]
\centering
  \includegraphics[width=130mm,scale=0.5]{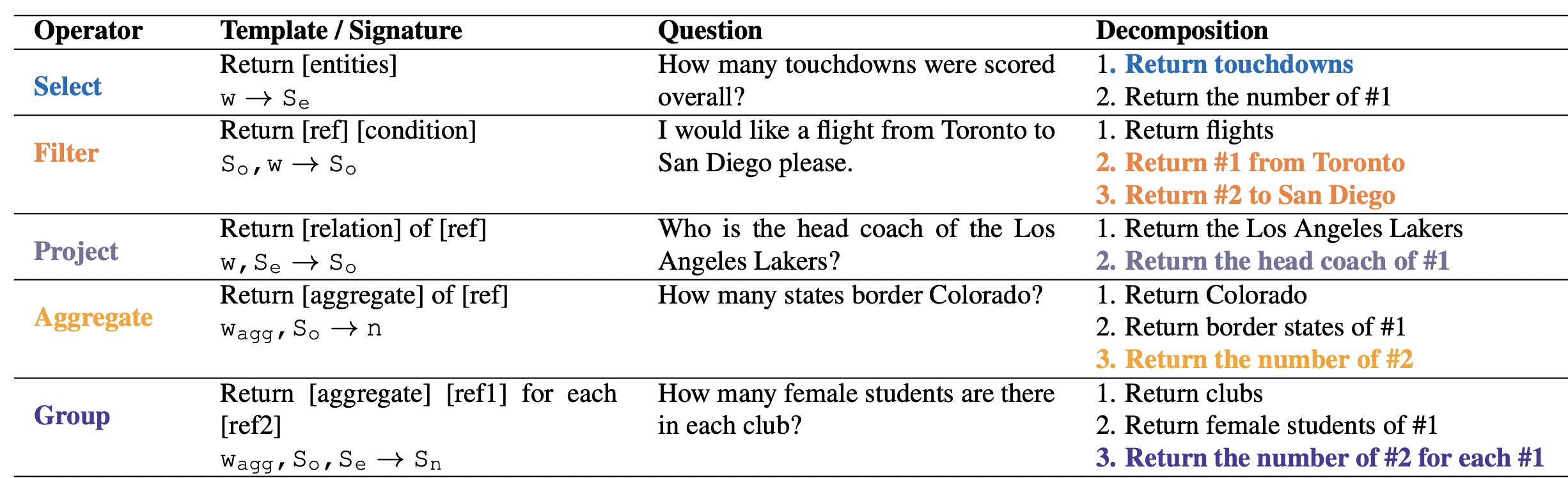}
  \caption{BREAK}
  \label{fig:break} 
\end{figure}

Many of methods for question answering like \citep{Yavuz2022} are not using semantic parsing. One reason is because current semantic parsing formalisms are expensive from dataset development perspective and also hard to implement. Question decompositon research is rapidly growing as is mentioned in \citep{Min2019},\citep{Perez2020}. 
A good approach to distant supervision in semantic parsing is to use backpropagation of errors that are generated from the gold solution in Figure~\ref{fig:questions}. Thus the semantic logical rules in each subproblem in this question decomposition are considered as latent variables and are not directly involved in supervision.
Question decomposition is so inspiring that \citep{Wolfson2020} introduced BREAK dataset and defined QDMR(Question Decomposition Meaning Representation) and contains over 83K pairs of questions and their QDMRs which can also be used for open domain question answering. BREAK dataset has thirteen operators and five of them is shown in Figure~\ref{fig:break}. By leveraging CopyNet in \citep{Gu2016} for BREAK dataset, this semantic parsing problem can be seen as a machine translation problem. Although problem seems to be solved with more than 70 percent accuracy, but this approach to modeling lacks interpretability and therefore  compositionality is necessary for better generalization.

\section{Model\label{model}\label{model}}
The motivation of using POS tags instead of the word tokens is reducing the complexity of the model. There is an even better representation that POS which is called "universal semantic tags" in \citep{Abzianidze2017}  as it includes semantic virtues of POS-tags and Named Entity (NE) classes but is not used in the present paper.

Using word2vec for words of the sentence would assign a high dimensional vector to each word while assigning a small size vector to each POS tag expresses that tag sufficiently and there is no need to represent words by vectors which reduces the curse of dimensionality. As will be shown in the experiments in the next section, It suffices that two or three dimensional vector for each tag capture an expressive representation.
POS tags embeddings are learnt jointly with semantic operators. Thus, this would lead to a good tradeoff for model complexity to have less prediction error and also avoiding overfitting. 

\subsection{Model Overview}
The proposed model is just like a standard encoder decoder network with attention like \citep{Bahdanau2014}. $x_t$ is the input POS tag sequence of a sentence, and $y_t$ is the resulting semantic operator sequence. Gated recurrent units(GRU) is used for both encoder and decoder since they are relatively faster than LSTM and they have less parameters. 
The hidden vectors of the encoder are:
\begin{equation}
h(t)=f(x_t,h_{t-1})
\end{equation}
The probability of each semantic operator sequence is :
\begin{equation}\label{eq-prob-operator}
\begin{split}
p(y)&=\prod_{t=1}^{T}p(y_{t}|y_1,\ldots,y_{t-1},x) \\
p(y_{t}|y_1,\ldots,y_{t-1},x)&=g(y_{t-1},s_t,c_t)
\end{split}
\end{equation}
where $s_t$ is the hidden state of the decoder at time t and $c_t$ is the context at time t. 
The weights of Bahdanau attention \citep{Bahdanau2014} is used to attend to different POS tags to align semantic operators with POS tags. Thus, the context vector $c_t$ is written as:
\begin{equation}
c_{t}=\sum_{j=1}^{T_x}\alpha_{tj}h_j
\end{equation}
where the weights $\alpha_{tj}$ are as follows:
\begin{equation}
\alpha_{tj}=\frac{\exp (a(s_{t-1},h_j))}{\sum_{k=1}^{T_{x}}\exp (a(s_{t-1},h_k))}
\end{equation}
The alignment model is the following single layer perceptron:
\begin{equation}
a(s_{t-1},h_{j})=v_{a}^{T} \tanh (W_{a}s_{t-1}+U_{a}h_{j})
\end{equation}
where  $v_{a},W_{a},U_{a}$ are weights that should be trained.
 Finally, the function g in equation~\ref{eq-prob-operator} is used to predict the semantic operators and is simply models as a linear layer with weights $W_{op}$ acting on concatenation of previous predicted semantic operator, context at time t, and previous hidden state of decoder. Thus:
 \begin{equation}
 g(y_{t-1}, s_t, c_t)=W_{op} (y_{t-1};c_{t};s_{t-1})
 \end{equation}

\subsection{Experiments}

\begin{figure} [h!]
\centering
  \includegraphics[width=100mm,scale=0.5]{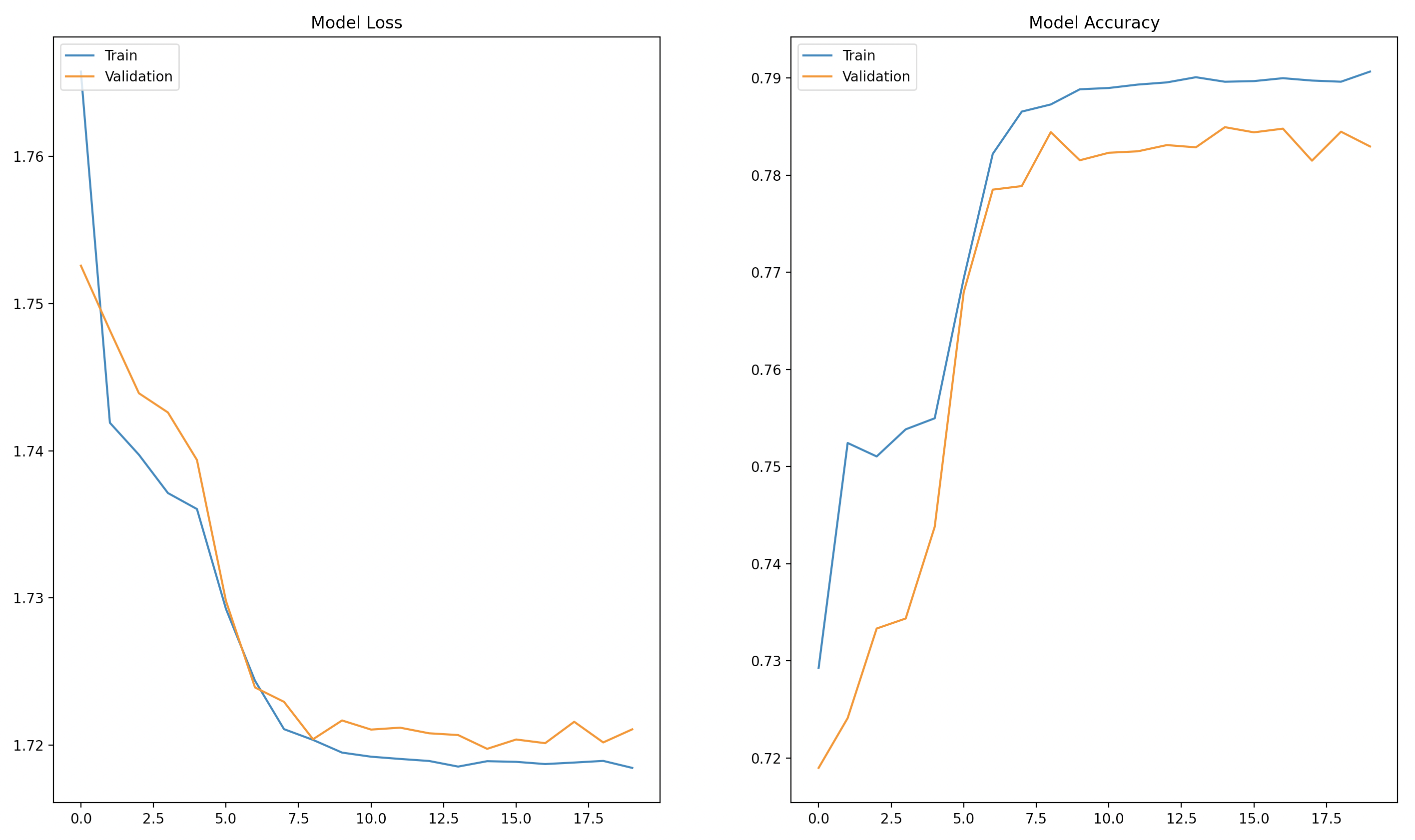}
  \caption{experiment 1}
  \label{fig:ex1} 
\end{figure}

\begin{figure} [h!]
\centering
  \includegraphics[width=100mm,scale=0.5]{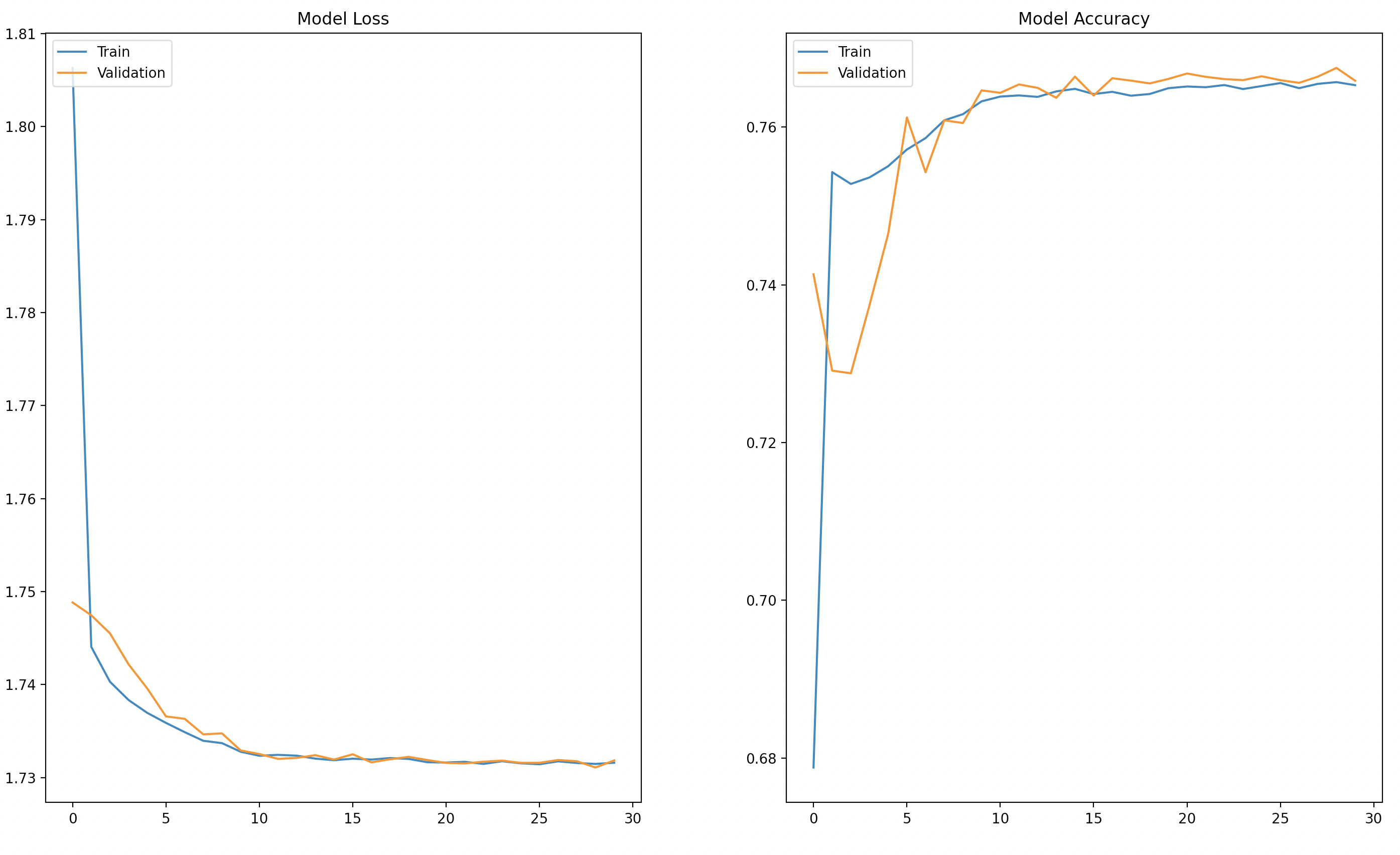}
  \caption{experiment 2}
  \label{fig:ex2} 
\end{figure}

Figure \ref{fig:ex1} shows the result of the first experiment. In this case, the embedding dimension as well as hidden state dimension of the encoder is kept as small as 3 since there is no need to increase the complexity of the model. Teacher forcing has been used to accelerate the speed of training. The dimension for both encoder and decoder embedding and hidden states in the second experiment in figure \ref{fig:ex2} is increased, but no significant improvement in accuracy has been observed, which once again reveals that POS tags do not need a big size vector to be represented. There are just 13 semantic operators and therefore the complexity of embedding and hidden states of the decoder is also kept small to reduce model complexity. A scheduler is used in both experiments to reduce the learning rate every 10 epochs. A pretrained word2vec model for POS tags could have been used but the present paper learns the embeddings of both POS tags and semantic operators along with the model jointly.

\begin{table}[h!]
  \begin{center}
    \caption{Experiments}
    \label{tab:table1}
	\begin{tblr}{h{3em}h{3em}h{3em}h{3em}h{3em}h{3em}h{3em}h{3em}h{3em}h{3em}}
	\hline
	{ex} & {opt} & {epochs} & {starting lr} & {batch size} & {teacher forcing} & {encoder embDim}& {encoder hidDim}& {decoder embDim}& {decoder hidDim} \\
      	\hline
      	1 & Adam & 20 & 1e-3 & 10 &  0.5 & 3 & 3 & 3 & 3\\
      	2 & SGD & 30 & 1e-2 & 5 &  0.5 & 5 & 10 & 4 & 12\\
    \end{tblr}
  \end{center}
\end{table}

\section{Applications}
Two applications of the model in section~\ref{model} is given.  The first application shows how semantic operator prediction could be used for a more expressive CopyNet Model by adding semantic operators as an extra feature. 
The second application shows how these operators could be used to have more expressive scores in the graph based approach which can be trained by max-margin loss or even a simple cross entropy loss as is used in \citep{Pasupat2019}.

\subsection{Conditioning To Enhance CopyNet}
Leveraging CopyNet idea \citep{Gu2016} for supervised learning of QDMR is straightforward and is done by many researchers.
The semantic operator prediction in the present paper can be used as an extra feature for CopyNet and It could be implemented in different ways. The original formulation of copyNet in \citep{Gu2016} uses the following probability to generate a target word $y_t$:
\begin{equation}
p(y_t|s_t,y_{t-1},c_{t},M)=p(y_{t},g|s_{t},y_{t-1},c_{t},M)+p(y_{t},c|s_{t},y_{t-1},c_{t},M) 
\end{equation}
where $M=\{h_1,\ldots,h_{T_S} \}$, g is the generator-mode, c is the copy-mode and $c_{t}$ is the context at time t. Now the result of the  model in section~\ref{model} could be used to condition on an extra expressive feature which is semantic operator $alsop_{t}$ at decoder time step t. Thus,
\begin{equation}
p(y_t|s_t,y_{t-1},c_{t},M,alsop_{t})=p(y_{t},g|s_{t},y_{t-1},c_{t},M,alsop_{t})+p(y_{t},c|s_{t},y_{t-1},c_{t},M,alsop_{t}) 
\end{equation}
There are many ways to model $p(y_{t},g|s_{t},y_{t-1},c_{t},M,alsop_{t})$ but the new problem is how to align the semantic operator prediction called by $sop_{t'}$ with the decoder time steps to model $alsop_{t}$. Note that $t'$ in $sop_{t'}$ refers to decoder for operator prediction while t in $alsop_{t}$ refers to time step of the decoder in CopyNet model and they should be aligned.
One idea is to define two actions namely "use{\_}current" action and "use{\_}next". This can be modeled by a softmax function followed by multilayer perceptron(MLP) to predict these two labels. The first label "use{\_}current" informs the decoder to just use the current prediction of semantic operator and they are still aligned. The second label "use{\_}next" expresses the fact that a misalignment has occurred and it has to move the pointer one step forward to make both sequences align. Thus the following MLP is used to model it:
\begin{equation}\label{mlp}
action(t)=softmax(MLP(y_{t-1},sop_{t'}))
\end{equation}
where $y_{t-1}$ in equation~\ref{mlp} shows that the action is very sensitive to the words that are produced by the CoyNet decoder model. It is also sensitive to the value operator prediction at time step $t'$ of the latent model.
Now, $alsop_{t}$ is obtained by the  following relation:
\[
    alsop_{t+1} = \left\{\begin{array}{lr}
        sop_{t'}, & \text{if action is current }\\
        sop_{t'+1}, & \text{if action is next } 
        \end{array}\right\} 
  \] 

The simplest idea is adding a new loss coming conditioning also on $sop_t$ which is latent variable with value from 13 operators.
This can be imagined as an Expectation Maximization(EM) model that in the expectation step, the operator prediction model of the present paper is calculated and in the maximization step the parameter of CopyNet model are learnt. Training could be separated or end to end. Thus, the following negative likelihood should be minimized.
\begin{equation}
L_{enh}= -\frac{1}{N}\sum_{k=1}^{N}\sum_{t=1}^{T}\log [p(y_{t}^{(k)}|y^{(k)}_{<t},X^{(k)},alsop^{(k)}_{<t}) ]
\end{equation}
\subsection{Parsing graph and Scoring}
A new graphbank for QDMR could be constructed for supervised learning. After a graphbank is created based on this new formalism, graph scoring methods could be easily utilized for parsing.
\begin{figure} [h!]
\centering
  \includegraphics[width=100mm,scale=0.5]{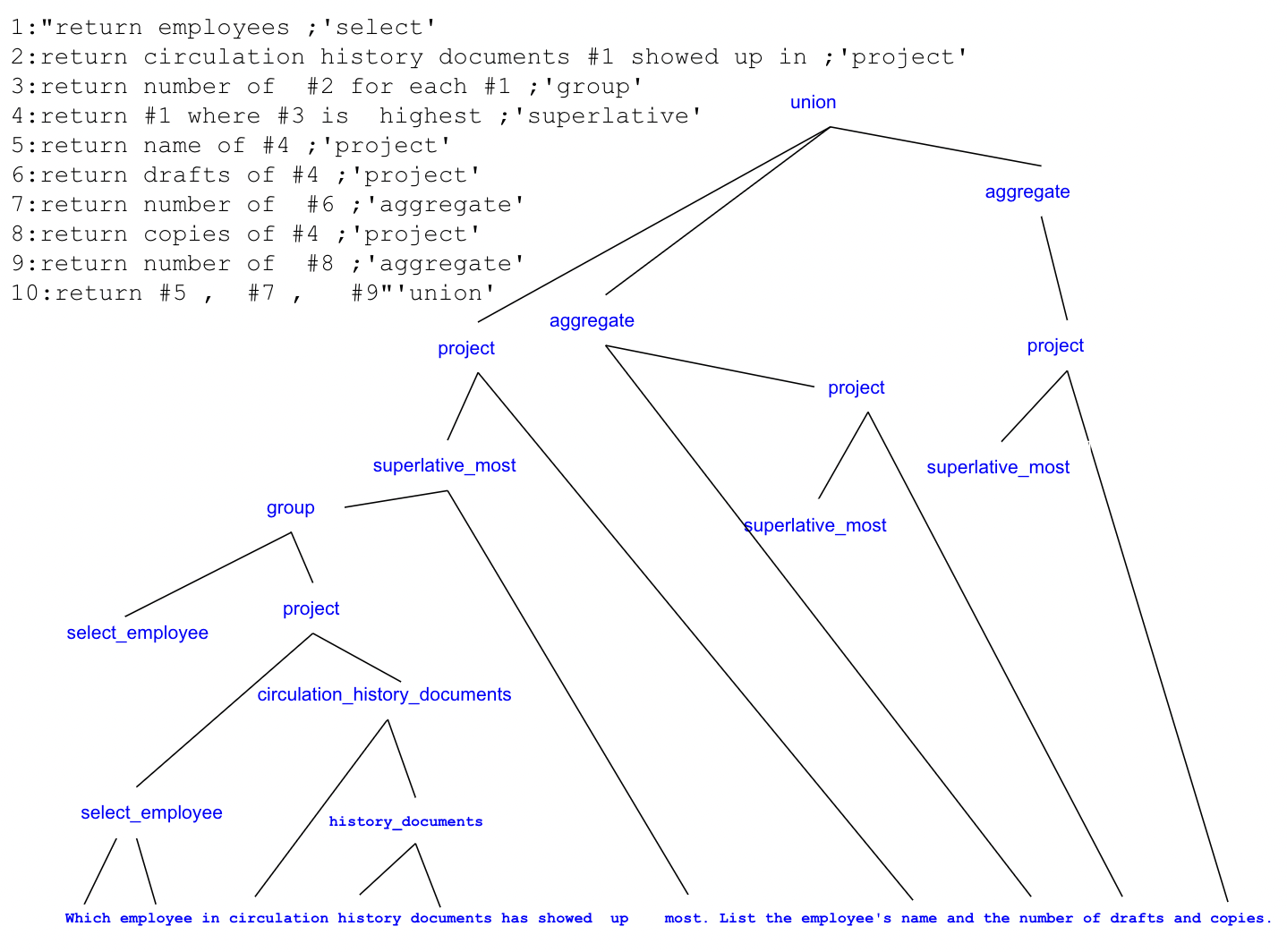}
  \caption{parsing graph}
  \label{fig:parsinggraph} 
\end{figure}

Figure~\ref{fig:parsinggraph} shows that the graph is not a tree and should be represented by a general graph instead. The standard way to treat this issue is to linearize the graph and turning it into a tree such that after calculating the scores of the nodes of the tree, the best tree will be obtained by the standard CKY algorithm. Although figure~\ref{fig:parsinggraph} is not a tree and is a directed acyclic graph(DAG) and therefore CKY algorithm can not be used directly, but hypergraph methods like \citep{Klein2001} or ideas based on dynamic extension of Dijkstra’s algorithm could be easily used for parsing.

The method of \citep{Socher2010} was first introduced for syntactic parsing but it could also be used for semantic parsing by some considerations. First consider a dataset of (sentence,tree) pairs denoted by $(x_{i},y_{i})$. Now, instead of syntactic tags in context free grammar(CFG), CCG tags(supertags) can be used. 
The total score of each tree can be computed as the sum of scores of each collapsing decision:
\begin{equation}
s(x_{i},y_{i})=\sum_{d\in T(y_{i})}s_{d}(c_{1},c_{2})
\end{equation}
where $s_{d}(c_{1},c_{2})$ is the score of each node of a tree
\begin{equation}
\begin{split}
s_{d}(c_{1},c_{2})&=W^{score}p \\
p&=\tanh (W[c_{1};c_{2}])+b^{(1)})
\end{split}
\end{equation}
Similar to score variable, another scalar variable can be defined which is called valence. Valence of a node(val) is defined as the number of all grammar rules below it. 
\begin{equation}
val_{d}(c_{1},c_{2})= W^{val}p
\end{equation}
The essence of valence is creating an order on all nodes. The terminal nodes have valence 0 and as it gets closer to the start symbol which is the root, the valence number increases. Thus root nodes have the highest valence number. The semantic operator prediction presented in the present paper can be expressed as a sequence of valance numbers which is a en expressive feature in the learning process whether the treebank is available(full supervision) or the size of treebank is too small(semisupervised case). 
The following objective should be maximized:
\begin{equation}
J=\sum_{i}s(x_{i},y_{i})-\underset{y\in A(x_{i})}\max (s(x_{i},y)+\Delta (y,y_{i}) )+ val(x_{i},y_{i})-\underset{y\in A(x_{i})}\max (val(x_{i},y)+\Delta (y,y_{i}) )
\end{equation}
where $\Delta$ is penalizing trees more when they deviate from the correct tree and has the following formula:
\begin{equation}
\Delta(y,y_{i})=\sum_{d\in T(y)} \mathds{1} \lambda d \not\in T(y_i)
\end{equation}

 \section{Conclusion}
 Different paradigms to solve semantic parsing problems is analyzed in the present paper which reveals the importance of distant supervision as well as those methods that create middle layer to reduce facing the problem directly at once. This makes the sentences to be understood by the machine in stages which would result a more scalable framework for semantic parsing.
 Finally, a fast method is presented for semantic operator prediction and the applications of it are demonstrated in different models. 
 \section{Future Work}
 One of the most important ideas to increase the accuracy of the present paper is to use "universal semantic tagging" which is introduced in \citep{Abzianidze2017} since POS tags used in the present paper fall short of providing sufficient information for lexical semantics. Thus, new categories are used in \citep{Abzianidze2017} to resolve this important issue by introducing 13 meta tags and 73 semantic tags. 
 For example the word "most" in Figure~\ref{fig:parsinggraph} could be represented by meta tag "COM" which stands for comparative and the semantic tag of "TOP". 
 
\bibliographystyle{agsm}
\bibliography{semanticoperator}
\end{document}